\documentclass[a4paper,fleqn]{cas-dc}

\usepackage[authoryear,longnamesfirst]{natbib}
\usepackage{hyperref}
\usepackage{multirow}
\usepackage{tabularx}
\usepackage{makecell}
\usepackage{algorithm2e}
\def\tsc#1{\csdef{#1}{\textsc{\lowercase{#1}}\xspace}}
\tsc{WGM}
\tsc{QE}
\tsc{EP}
\tsc{PMS}
\tsc{BEC}
\tsc{DE}

\begin{document}
\let\WriteBookmarks\relax
\def\floatpagepagefraction{1}
\def\textpagefraction{.001}

\shorttitle{Fine-grained Stateful Knowledge Exploration}

\shortauthors{Dehao Tao et~al.}

\title [mode = title]{Fine-grained Stateful Knowledge Exploration: Effective and Efficient Graph Retrieval with Large Language Models}                      
\tnotemark[1]

\tnotetext[1]{This document is the results of the research
   project funded by the National Key R\&D Program of China (2022ZD0115801)}


%
\author[1]{Dehao Tao}[style=chinese]

\credit{Conceptualization, Data curation, Formal analysis, Investigation, Methodology, Software, Validation, Writing – original draft, Writing – review and editing}



\affiliation[1]{organization={Tsinghua University},
    addressline={Haidian District}, 
    city={Beijing},
    postcode={100084}, 
    country={China}}

\author[1]{Congqi Wang}[style=chinese]
\credit{Software, Validation, Writing – review and editing}
\author[2]{Feng Huang}[style=chinese]
\credit{Software, Validation}
\author[1]{Junhao Chen}[style=chinese]
\credit{Writing – review and editing}
\author[1]{Yongfeng Huang}[style=chinese]
\credit{Funding acquisition, Writing – review and editing}
\cormark[1]
\author[1]{Minghu Jiang}[style=chinese]
\credit{Writing – review and editing}

\affiliation[2]{organization={Xinjiang University},
    addressline={No. 666, Shengli Road, Tianshan District},  
city={Urumqi},  
postcode={830046},  
country={China}  }

\cortext[cor1]{Corresponding author}



\begin{abstract}
Large Language Models (LLMs) have shown impressive capabilities, yet updating their knowledge remains a significant challenge, often leading to outdated or inaccurate responses. A proposed solution is the integration of external knowledge bases, such as knowledge graphs, with LLMs. Most existing methods use a paradigm that treats the whole question as the objective, with relevant knowledge being incrementally retrieved from the knowledge graph. However, this paradigm often leads to a granularity mismatch between the target question and the retrieved entities and relations. As a result, the information in the question cannot precisely correspond to the retrieved knowledge.  This may cause redundant exploration or omission of vital knowledge, thereby leading to enhanced computational consumption and reduced retrieval accuracy. To address the limitations of coarse-grained knowledge exploration, we propose FiSKE, a novel paradigm for \textbf{Fi}ne-grained \textbf{S}tateful \textbf{K}nowledge \textbf{E}xploration. FiSKE first decomposes questions into fine-grained clues, then employs an adaptive mapping strategy during knowledge exploration process to resolve ambiguity in clue-to-graph mappings. This strategy dynamically infers contextual correspondences while maintaining a stateful record of the mappings. A clue-driven termination mechanism ensures rigorous augmentation—leveraging fully mapped paths for LLMs while reverting to chain-of-thought reasoning when necessary. Our approach balances precision and efficiency. Experiments on multiple datasets revealed that our paradigm surpasses current advanced methods in knowledge retrieval while significantly reducing the average number of LLM invocations. The code for this paper can be found at \href{https://github.com/nnnoidea/stateful-KGQA.}{https://github.com/nnnoidea/stateful-KGQA.}

\end{abstract}


\begin{highlights}
\item First to systematically diagnose the granularity mismatch problem in mainstream KG-LLM paradigms, revealing its threefold impact: computational redundancy, incomplete retrieval, and unnecessary prior knowledge dependence.
\item Propose FiSKE, the first stateful fine-grained KG exploration framework, enabling atomic question decomposition, cross-iteration state tracking, and dynamic switching between KG augmentation and chain-of-thought reasoning.
\item Resolve clue-to-KG mapping ambiguity via a novel adaptive strategy that dynamically infers contextual correspondences (entity/relation/both), eliminating redundant exploration while ensuring retrieval precision.
\end{highlights}


\maketitle

\section{Introduction}

Large language models (LLMs) have demonstrated considerable efficacy in a range of natural language processing tasks, as evidenced by recent studies \cite{achiam2023gpt,cheng2023gpt,meta2024introducing,zhao2024comi,li2024unigen}. Nevertheless, there are certain limitations to be considered when employing LLMs in question-answering tasks. The knowledge of LLMs remains static, fixed at the point of their training, which presents a challenge in incorporating timely updates. Moreover, in response to queries with which they are unfamiliar, LLMs may occasionally provide incorrect answers accompanied by an overconfident tone \cite{luo2023augmented}, a phenomenon that is often referred to as hallucination \cite{DBLP:journals/corr/abs-2302-04023,DBLP:journals/csur/JiLFYSXIBMF23}. In light of these challenges, an effective solution is the integration of external knowledge \cite{nie2020}, specifically knowledge graphs (KGs), to enhance the question-answering proficiency of LLMs.
\begin{figure}
    \centering
    \includegraphics[width=1\linewidth]{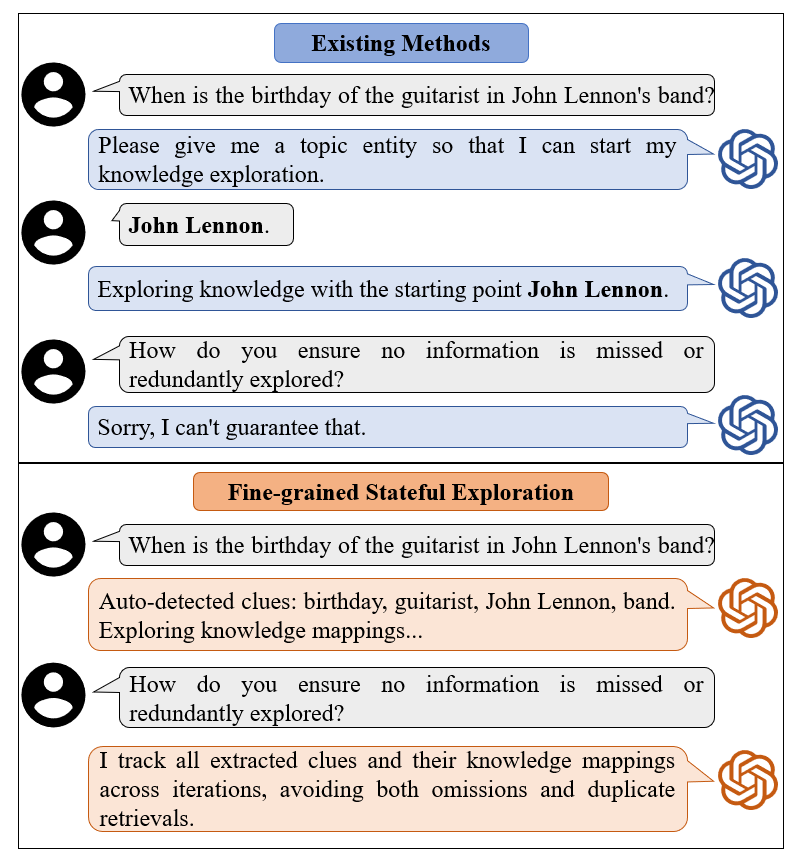}
    \caption{Compared to existing methods, our approach exhibits two key distinctions: (1) no manual provision of topic entities is required, and (2) it eliminates both information oversight and redundant exploration.}
    \label{fig:qa}
\end{figure}

A number of studies have been conducted with the objective of combining external knowledge with LLMs. One straightforward approach is to prompt LLMs to directly generate the corresponding knowledge base operation statements based on questions. This method has been demonstrated to be effective in various studies, as referenced in \cite{hu2023chatdb,wang2023knowledgpt}. Additionally, chain-of-thought \cite{wei2022chain} is employed to assist LLMs in decomposing input questions into multiple sub-questions \cite{li2023chain,DBLP:conf/acl/ZhaoLJQB23}. However, these methods do not take into account the awareness of the nature of external knowledge sources on the part of LLMs. Consequently, the instructions or sub-questions generated by LLMs may not be aligned with the external knowledge. 
To avoid the mismatch issue due to unawareness of the knowledge base, ToG \cite{sun2024think} leverages the semantic capabilities of LLMs to retrieve relevant knowledge from the knowledge graph. 

However, existing methods primarily focus on addressing mismatches at the character level, while largely overlooking the more critical issue of mismatches at the information granularity level. A question, as a sentence comprising multiple pieces of information, inherently exhibits a coarser information granularity compared to the fine-grained entities and relations in the knowledge graph. As shown in the upper half of Fig. \ref{fig:qa}, this discrepancy gives rise to the following two major challenges. 

First, existing methods can only verify the relevance of retrieved knowledge (entities/relations) to the question, but not their redundancy or completeness. Consequently, current approaches are constrained to stateless iterative exploration (i.e., processing each iteration independently without retaining cross-iteration information), wherein the identical sentence must be repeatedly reprocessed across iterations, each time realigning with newly retrieved subgraphs containing different entity-relation combinations. This stateless nature inherently leads to two significant inefficiencies: redundant exploration and information omission. Specifically, certain pieces of information in the sentence may be redundantly explored across different rounds, while others may remain entirely overlooked throughout the process. Such limitations not only compromise the accuracy of knowledge retrieval but also result in suboptimal computational efficiency. Notably, while some approaches attempt to mitigate granularity mismatch through query splitting \cite{chen2024plan}, these patchwork solutions fail to address the root cause: the persistent disparity between question-level abstraction and KG-level granularity. Second, current methodologies require manual provision of fine-grained topic entities as exploration starting points \cite{sun2024think,chen2024plan}, inevitably demanding additional human effort in practical applications. These dual challenges highlight the need for a state-aware approach capable of dynamically bridging question contexts with KG knowledge.

\begin{figure*}
    \centering
    \includegraphics[width=1\linewidth]{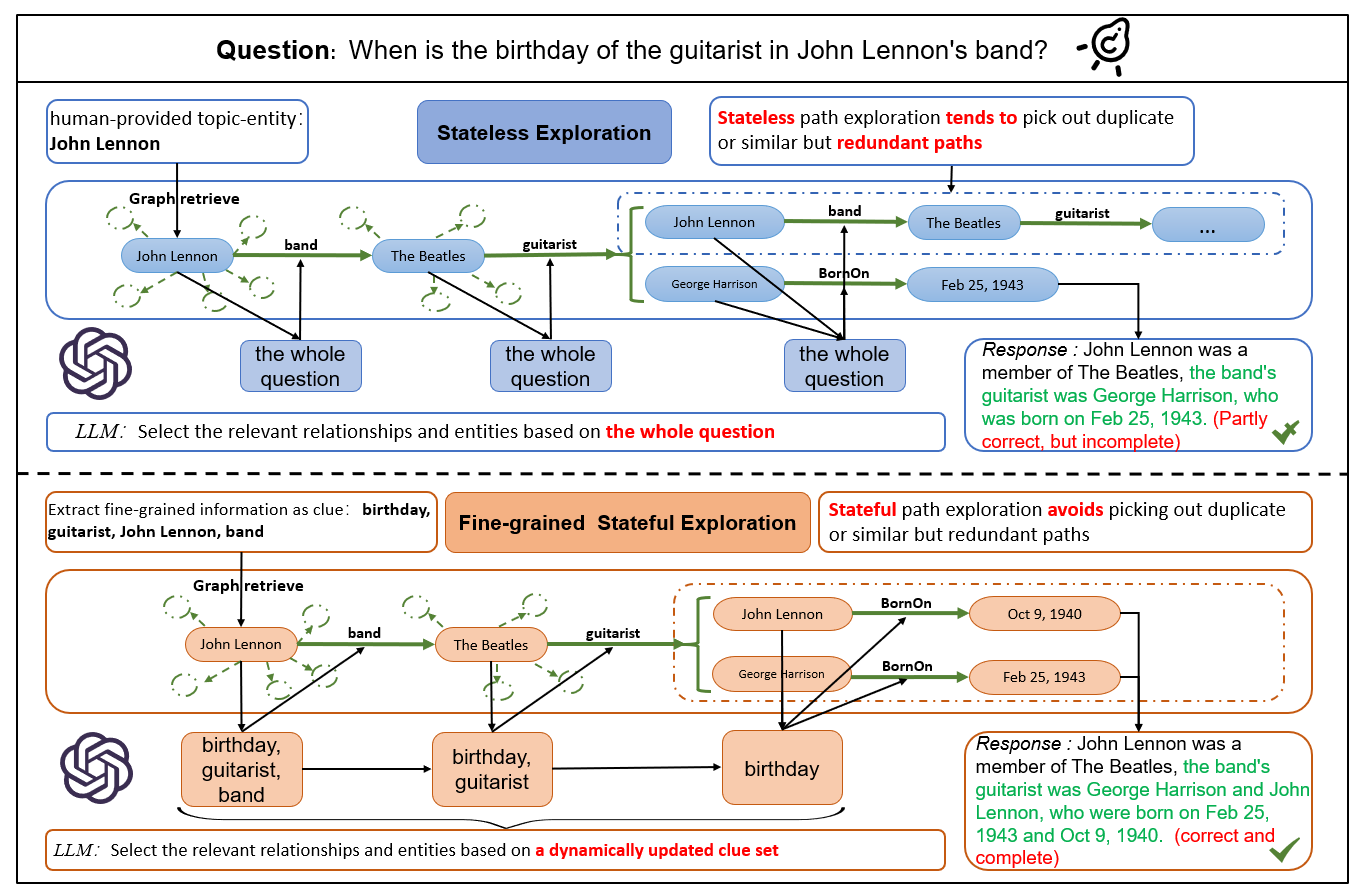}
    \caption{A simple example illustrating the differences between the paradigm we propose and the existing main stream paradigm. Stateless exploration may perform redundant exploration at times and may miss information. In contrast, the stateful exploration we propose can avoid redundancy and omissions.}
    \label{fig:comparison}
\end{figure*}

To solve the above problems caused by the mismatch of information granularity, in this paper we propose a novel paradigm of \textbf{Fi}ne-grained \textbf{S}tateful \textbf{K}nowledge \textbf{E}xploration, which we refer to as FiSKE, as illustrated in the lower part of Fig. \ref{fig:qa}. This new paradigm extracts fine-grained, sufficient information from the question to capture most of its semantics. It also shifts the goal of knowledge exploration from coarse-grained to multiple independent fine-grained information units, which we refer to as clues. During knowledge exploration, the knowledge corresponding to each clue is iteratively retrieved from the knowledge graph. We define this correspondence between the clues and the knowledge in the graph as a mapping. By maintaining a stateful record of the clues and their mappings across iterations, FiSKE ensures that different rounds of exploration avoid redundant information retrieval while preventing the oversight of critical details. With its clue-driven termination mechanism, FiSKE employs only fully mapped knowledge paths for LLM augmentation. If no valid paths are found, it automatically falls back to chain-of-thought reasoning. This hybrid approach ensures both rigor and efficiency in knowledge utilization. A simple path exploration example is shown in the Fig. \ref{fig:comparison}, demonstrating the impact of our proposed stateful method on avoiding redundant exploration.

However, this fine-grained paradigm introduces a unique challenge in mapping resolution: the extracted clues may correspond to different structural elements in the knowledge graph. Specifically, while some clues distinctly map to either entities or relationships, others may simultaneously satisfy the conditions for both. This ambiguity arises precisely because our fine-grained method is designed to evaluate the redundancy or completeness of retrieved knowledge, a capability absent in existing approaches. Current methods rely on coarse question-entity mapping, which lacks mechanisms to assess knowledge redundancy/completeness, let alone resolve such ambiguous mappings. To address this, we introduce an adaptive mapping strategy that dynamically resolves whether a clue corresponds to an entity, a relationship, or both, based on contextual evidence from the knowledge graph. 

The main contributions of this paper are listed as follows:
\begin{itemize}
    \item We identify and highlight the critical issue of information granularity mismatch in existing mainstream paradigms. This mismatch forces current methods to rely on stateless iterative exploration, leading to increased computational overhead during knowledge exploration, incomplete retrieval of knowledge, and unnecessary reliance on prior knowledge. Our analysis provides a deeper understanding of the limitations inherent in these approaches.
    \item We propose FiSKE, a novel paradigm that transforms knowledge exploration into a stateful, fine-grained process. By decomposing questions into atomic clues and maintaining cross-iteration exploration states, FiSKE eliminates redundancy and omissions inherent in stateless approaches. The paradigm further integrates a clue-driven termination mechanism to dynamically balance KG augmentation and chain-of-thought reasoning.
    \item We identify a key challenge in fine-grained knowledge exploration: extracted clues may have ambiguous mappings to knowledge graph elements, potentially corresponding to entities, relations, or both. To resolve this mapping ambiguity, we propose an adaptive mapping strategy that dynamically determines the most appropriate mapping for each clue based on contextual evidence from the knowledge graph.
    \item We conduct extensive experiments on multiple datasets to validate the effectiveness of our proposed paradigm. The results demonstrate that our method achieves superior performance across all datasets, outperforming existing methods in knowledge graph question-answering tasks. Additionally, our paradigm drastically reduces the number of large language model calls, often reducing them to a fraction of those required by current approaches, thereby improving computational efficiency.
\end{itemize}

\section{Related work}

\subsection{Document Retrieval Enhanced Models}
One direct approach to addressing the generation of illusions by LLMs is the use of external knowledge \cite{DBLP:journals/corr/abs-2302-07842} to enhance the model's ability to perceive the real world. This has led to the emergence of numerous knowledge-based models. Information retrieval is commonly employed in developing knowledge-based chatbots \cite{DBLP:conf/emnlp/0001PCKW21}. BlenderBot2 \cite{xu2021goldfish} integrates Internet search capabilities. SeeKeR \cite{DBLP:conf/emnlp/0001KARSW22} surpasses BlenderBot2 \cite{xu2021goldfish} by executing three modular tasks using a single language model: generating search queries, extracting relevant knowledge from retrieved documents, and generating the final response. BlenderBot3 \cite{DBLP:journals/corr/abs-2208-03188} fine-tunes a 175-billion-parameter OPT \cite{DBLP:journals/corr/abs-2205-01068} on a combination of 20 question-answering and dialogue datasets. Atlas \cite{DBLP:journals/jmlr/IzacardLLHPSDJRG23} stands out as a top model on the KILT benchmark \cite{DBLP:conf/naacl/PetroniPFLYCTJK21}, which includes 11 knowledge-oriented tasks, including Wizard of Wikipedia \cite{DBLP:conf/iclr/DinanRSFAW19}.

\subsection{Knowledge Graph Enhanced Models}
There are diverse ways to enhance models based on knowledge, aiming to address issues arising from complex reasoning, hallucination, and poor interpretability. Cao et al \cite{DBLP:conf/acl/CaoZ000CP20} improved implicit event causality extraction by leveraging structured knowledge bases. Zhao et al \cite{DBLP:conf/acl/ZhaoLJQB23} recently proposed a Verification and Editing (VE) framework. This framework improves the Chain of Thought (CoT) reasoning in large language models (LLMs) by integrating retrieval systems \cite{wei2022chain}.
Researchers have also introduced the Chain-of-Knowledge (CoK) \cite{li2023chainofknowledge}, a novel framework designed to enhance the factual correctness of LLMs using heterogeneous knowledge sources. This ensures that inaccuracies in the foundational principles do not propagate to subsequent steps. Think-on-Graph(ToG) \cite{sun2024think} is a plug-and-play framework that seamlessly applies to various LLMs and KGs. ToG also enhances the reasoning capabilities, enabling smaller LLMs to compete with larger LLMs.

\section{Methodology}
\label{}

\subsection{Problem Definition}

Knowledge Graph Question-Answering is the task of retrieving relevant knowledge from a knowledge graph to answer given questions. In our proposed new paradigm, we first obtain a set of clues \(clue_{all} = \{\text{clue}_1, \text{clue}_2, \dots \}\) from the question q. For each clue in the \(clue_{all}\), we explore its corresponding relations \(r\) or entities \(e\) in the knowledge graph, forming a knowledge path \(P\) = \(\{(e^{0}, r^1, e^1), \dots,(e^{i-1}, r^i, e^i)\}\). This path then serves as a base for the large language model to answer the question.

\subsection{Extraction of Fine-grained Clues}
\label{fine}
To enable stateful knowledge exploration, it is essential to extract fine-grained information, which we refer to as clues. We posit that the majority of information within a question can be effectively characterized by its keywords, whose granularity is comparable to that of entities in the knowledge graph. Using these keywords as clues facilitates knowledge retrieval and enables the resolution of the given question.

For example, in the question ``What is the population of the capital of China?", the key information required is ``population", while ``China" and ``capital" provide contextual constraints. Thus, the keywords extracted from this question are (China, capital, population), which serve as the clues for knowledge exploration.

Clues may correspond to entities, relations, or their combinations.
We leverage the capabilities of the LLM to systematically extract the complete set of clues, resulting in an ontology-agnostic approach. Example prompt can be seen in Appendix \ref{appendix prompt}. The exact mappings are resolved dynamically during path exploration through our adaptive mapping strategy.

\subsection{Starting Point}
\label{starting point}
After extracting the fine-grained clues, our approach uses these clues to identify exploration starting points in the knowledge graph. Unlike existing methods that require human-provided topic entities, we automatically determine starting points through exact lexical matching (e.g., using a basic database operation like SELECT or MATCH). Since well-formed questions typically contain at least one identifiable entity reference, this matching process reliably anchors the exploration. Each successfully matched entity serves as a starting point, from which we iteratively explore the complete knowledge path.

\subsection{Adaptive and Stateful Knowledge Exploration}
\label{ke}
Through iterative exploration, we map clues to corresponding entities or relations, constructing a knowledge path $P$. We maintain a stateful record of these mappings to dynamically adjust candidate clues in each round. An adaptive mapping strategy determines whether current-step triples should map to single or multiple clues.

The strategy comprises two steps: relation-based and entity-based mapping exploration. Using the $n$-th iteration as an example, previous iterations yield $n-1$ triples forming path $P = \{(e^{0}, r^1, e^1), \dots,(e^{n-2}, r^{n-1}, e^{n-1})\}$, where $e^0$ is the initial head entity. The explored clue set $C$ prevents redundant processing, with candidate clues $C' = clue_{all} - C$.

\paragraph{Relation-Based Mapping Exploration}
In the $n$-th iteration, we first retrieve relations $R = \{r_1, \dots,r_m\}$ for entity $e^{n-1}$, then prompt LLM to map $C'$ to $R$, yielding:
$$
\text{r-mapping} = \{(\text{clue}_i, r_j)\}
$$
The process terminates if $\text{r-mapping} = \emptyset$.

\paragraph{Entity-Based Mapping Exploration}
For $\text{r-mapping} = [(\text{clue}_i, r_j)]$, we retrieve connected nodes $E = \{r_j: E_j\}$ from KG and prompt LLM to map $C'$ to $E$, yielding:
$$
\text{e-mapping} = \{(\text{clue}_k, e_l)\}
$$

Three outcomes occur:

\textbf{Case 1}: Single-clue dual mapping $(\text{clue}_1, r_i)$ and $(\text{clue}_1, e_j)$:
\begin{itemize}
\item Update $C \leftarrow C \cup \{clue_1\}$
\item Extend $P \leftarrow P \cup \{(e^{n-1}, r_i, e_j)\}$
\end{itemize}

\textbf{Case 2}: Dual-clue mapping $(\text{clue}_1, r_i)$ and $(\text{clue}_2, e_j)$:
\begin{itemize}
\item Update $C \leftarrow C \cup \{clue_1, clue_2\}$
\item Extend $P$ as in Case 1
\end{itemize}

\textbf{Case 3}: Terminate if $\text{e-mapping} = \emptyset$.

The strategy iteratively constructs $P$ through LLM mapping with stateful tracking of $C$ and $C'$. Example prompts appear in Appendix \ref{appendix prompt}. The pseudocode can be found in Appendix \ref{pseudocode}.

\subsection{Clue-Driven Termination and Branch-Adaptive Answering}
\label{answering}
During the exploration process, each step may generate multiple outcomes, resulting in multiple branches. To prevent an exponential increase in the number of branches, existing methods typically select the top-N branches as final results. However, this approach inevitably risks omissions, especially when the number of correct results exceeds N. In our method, we do not filter branches based on quantity but instead rely on whether all clues can be successfully mapped. Specifically, for any branch, if it fails during the mapping exploration phase, we deem it incapable of fully resolving all clues and thus discard it from further consideration. Conversely, if a branch successfully maps all clues, the exploration is considered complete. Upon finishing the current round of exploration across all branches, the process terminates, and no further exploration is conducted.

If a branch successfully completes all mappings, we provide its knowledge path \(P\) as external knowledge to the LLM, instructing the model to answer questions based on this contextual information. However, if all branches fail to achieve complete mapping, we infer that the retrieved knowledge paths \(P\) may contain incorrect or irrelevant information. To prevent such noisy knowledge from interfering with the LLM’s reasoning, we refrain from supplying P to the model. Instead, we employ chain-of-thought prompting to guide the LLM in leveraging its internal knowledge to generate answers. Example prompt can be seen in Appendix \ref{appendix prompt}.

\section{Experiments}
\subsection{Experimental Settings}

\textbf{Datasets and Evaluation Metrics.} To evaluate the performance of our proposed paradigm, We selected two open-source knowledge graphs and one self-constructed graph as external knowledge bases for experimentation. The open-source knowledge graphs are MOOC Q\&A \cite{yu2020mooccube} and Freebase \cite{bollacker2008freebase}. The graph that we constructed ourselves is not yet publicly available and is referred to in this paper as the agricultural knowledge graph.

MOOC Q\&A is a dataset collected from MOOC websites, comprising a knowledge graph and associated question-answer sets. The knowledge graph includes 7 categories of entities, totaling 15,388, with 11 types of relationships forming 277,487 triplets. The question-answer sets contain True/False Question and Query Question.

Freebase is a large-scale, semi-structured database supported by Google, designed to collect and connect information about millions of entities and their relationships worldwide. It covers a wide range of topics such as people, places, books, movies, music, etc., with each entity having a unique identifier. We conduct experiments on two QA sets with Freebase as external knowledge base: WebQSP \cite{yih2016value} and WebQuestions \cite{berant2013semantic}.  

The agricultural knowledge graph that we constructed includes over 100,000 entities and 1 million triples, on which basis we built single-hop and multi-hop question-answering sets.

For questions with multiple correct answers, we evaluate the results using two metrics: partial match and complete match. Partial match requires the experimental method to provide one or more correct answers, while complete match demands all correct answers.

\textbf{Baselines.} We compared our approach with six baseline methods: standard prompting (IO prompt) \cite{brown2020language}, Self-Consistency (SC) \cite{sc2023}, chain of thought prompting (CoT prompt) \cite{wei2022chain}, ToG \cite{sun2024think}, PoG \cite{chen2024plan}, StructGPT \cite{jiang2023structgpt} and KB-BINDER \cite{li2023few}. IO prompt, SC and CoT prompt are knowledge-free methods, used to measure how many questions LLMs can answer solely based on their internal knowledge. ToG, PoG, StructGPT and KB-BINDER represent previous state-of-the-art approaches in knowledge base question answering, serving as baselines to evaluate our knowledge retrieval method's effectiveness.

\begin{table*}[htb]
    \centering
    \caption{Results for MOOC Q\&A. We report the accuracy of each method in answering "yes" or "no" correctly on the True/False question set, as well as the proportion of responses containing partially correct answers (partial match) and fully correct answers (complete match) on the Query question set.}
    \resizebox{1\linewidth}{!}{
    \begin{tabular}{l|c|c|cc}
        \hline
         \multicolumn{1}{c|}{\multirow{2}{*}{\textbf{Method}}} & \multirow{2}{*}{\textbf{1-Hop True/False}}& \multirow{2}{*}{\textbf{Multi-Hop True/False}} & \multicolumn{2}{c}{\textbf{1-Hop Query}} \\
        \cline{4-5}
         & &  &\textbf{partial match}    & \textbf{complete match}   \\
        \hline
        \multicolumn{5}{c}{\textit{without external knowledge}}  \\
        \hline
        IO prompt \cite{brown2020language} w/GPT-3.5       & 8.0 &7.3 &10.0 & 7.5\\
        CoT \cite{wei2022chain} w/GPT-3.5      &3.4 &2.2 &10.0 &6.7 \\
        \hline
        \multicolumn{5}{c}{\textit{with external knowledge}} \\
        \hline
        ToG \cite{sun2024think} w/GLM4-9B-Chat & 48.7 &37.6 & 24.5& 8.2\\
        ToG \cite{sun2024think} w/GPT-3.5     & 53.8 &39.2 & 65.6 &46.9 \\
        ToG \cite{sun2024think} w/GPT-4& 54.3& 41.7&82.6 & 56.5\\
        \hline
        PoG \cite{chen2024plan} w/GLM4-9B-Chat & 57.0 & 44.6 & 52.3 & 43.3\\
        PoG \cite{chen2024plan} w/GPT-3.5 & 67.5 & 52.9 & 73.9 & 54.1 \\
        PoG \cite{chen2024plan} w/GPT-4& 77.5 & 62.9 & 82.9 & 64.5 \\
       
        \hline
        FiSKE(Ours) w/GLM4-9B-Chat & 84.1& 55.0 & 77.1 & 67.9 \\
        FiSKE(Ours) w/GPT-3.5      & 89.1 & 71.1 &83.1 &76.4 \\
        FiSKE(Ours) w/GPT-4 & \textbf{94.4} & \textbf{74.4} & \textbf{84.2} & \textbf{77.8} \\
        \hline
    \end{tabular}
    }
    \label{table:main}
\end{table*}

\textbf{Experiment Details.} To ensure the reliability and reproducibility of the experiments, we set the temperature parameter to 0 for all LLMs. The two hyperparameters of ToG are set to their default values. We modified all prompts for all baseline methods to be suitable for MOOC Q\&A and the agricultural knowledge grpah. During the experiment, we observed that ToG would invoke LLM hundreds of times to answer certain questions, exceeding ten minutes. To mitigate the impact of these extreme cases, we terminated the exploration for all methods when it exceeded 30 LLM calls. We conduct experiments on one thousand question-answer pairs from each dataset.

\begin{table*}[htb]
    \centering
\caption{Results for WebQuestions and WebQSP. To better illustrate whether these methods answer questions based on the inherent knowledge of LLMs or the retrieved knowledge, we merged the responses of these methods with the IO answers of LLMs and conducted additional comparisons. These are respectively represented as ToG + IO, PoG + IO and FiSKE + IO. }
\resizebox{1\linewidth}{!}{
\begin{tabular}{l|cc|cc}
        \hline
       \multicolumn{1}{c|}{\multirow{2}{*}{\textbf{Method}}}& \multicolumn{2}{c}{\textbf{WebQuestions}} & \multicolumn{2}{c}{\textbf{WebQSP}}    \\
        \cline{2-5}
        & \textbf{partial match}    & \textbf{complete match}     & \textbf{partial match}      & \textbf{complete match}   \\
        \hline
        \multicolumn{5}{c}{\textit{without external knowledge}} \\
        \hline
IO prompt \cite{brown2020language} w/Llama3-8B-Instruct &  44.4 & 37.2    &  52.4  &  27.0       \\ 
IO prompt \cite{brown2020language} w/GLM4-9B-Chat    &   52.4  &  33.4    &    60.1  &   31.8  \\ 
SC \cite{sc2023} w/Llama3-8B-Instruct  &  45.2   &   38.8   &   52.1     &  26.7     \\ 
SC \cite{sc2023} w/GLM4-9B-Chat   &   54.4  &   34.9   &  59.6     &    32.1     \\ 
COT \cite{wei2022chain} w/Llama3-8B-Instruct  &  54.3   &   35.8   &   62.1     &  33.7     \\ 
COT \cite{wei2022chain} w/GLM4-9B-Chat   &   54.8  &   34.3   &  63.0     &    32.5     \\ 

        \hline
        \multicolumn{5}{c}{\textit{with external knowledge}} \\
        \hline
ToG \cite{sun2024think} w/Llama3-8B-Instruct  &  55.1   &   36.7   &  55.6    &   32.3    \\
ToG \cite{sun2024think} w/GLM4-9B-Chat    &   63.3  &   37.8   &    63.3     &    30.6   \\
ToG w/GPT-3.5  &   - &  -  &    76.2     &  -  \\
ToG + IO w/Llama3-8B-Instruct &  67.3 &   43.9   &  69.7   &  39.4       \\
ToG + IO w/GLM4-9B-Chat  &   71.4 &   46.9   &    74.0     &   38.8   \\

\hline
PoG \cite{chen2024plan} w/Llama3-8B-Instruct     &   62.3  &   39.4   &    60.2     &    35.2   \\
PoG \cite{chen2024plan} w/GLM4-9B-Chat    &   63.4  &   38.3   &    63.4     &    34.4  \\
PoG w/GPT-3.5  &   - &  -  &    82.0     &  -  \\
PoG + IO w/Llama3-8B-Instruct &  70.9  &   42.1   &  70.8   &  40.3       \\
PoG + IO w/GLM4-9B-Chat  &   73.6 &   45.2   &    74.4     &   41.6   \\

\hline

StructGPT \cite{jiang2023structgpt} w/GPT-3.5     &  -  &   -   &    72.6     &    -   \\

KB-BINDER \cite{li2023few} w/GPT-3.5     &  -  &   -   &    74.4     &    -   \\

\hline

FiSKE w/Llama3-8B-Instruct &    70.3 &  44.6    &    70.8     &  40.4     \\
FiSKE w/GLM4-9B-Chat   &   66.3  &   38.8   &  62.5    &    36.4     \\
FiSKE w/GPT-3.5  &   - &  -  &   \textbf{83.1}    &  \textbf{49.1}  \\
FiSKE + IO w/Llama3-8B-Instruct  &   \textbf{81.2}  &  \textbf{50.5}  &   75.0   & 43.3     \\
FiSKE + IO w/GLM4-9B-Chat  &   77.9  &   50.2   &  73.9   &   43.2    \\   

        \hline
\end{tabular}
}
\label{table:freebase}
\end{table*}

\begin{table*}[htb]

    \centering
    \caption{Results for the agricultural graph. The knowledge in this graph and the corresponding question-answering sets are relatively specialized.}
    \resizebox{1\linewidth}{!}{
    \begin{tabular}{l|cc|cc}
        \hline
         \multicolumn{1}{c}{\multirow{2}{*}{\textbf{Method}}} & \multicolumn{2}{c}{\textbf{1-Hop Query}}& \multicolumn{2}{c}{\textbf{multi-Hop Query}} \\
        \cline{2-5}
         &  \textbf{partial match} & \textbf{complete match} & \textbf{partial match} & \textbf{complete match} \\
            \hline
        IO prompt \cite{brown2020language} w/GLM4-9B-Chat       & 14.3 &6.1 &6.1 & 2.0\\
        CoT \cite{wei2022chain} w/GLM4-9B-Chat    &3.4 &0.0 &4.1 &0.0 \\
        \hline
        ToG \cite{sun2024think} w/GLM4-9B-Chat &42.9 & 30.3& 10.0& 8.2\\
        FiSKE w/GLM4-9B-Chat   & 48.2 & 32.1 &9.8 & 9.3\\
        \hline
        
    \end{tabular}
    }
    
    \label{table:agr}
\end{table*}
\subsection{Performance Comparison}

We conducted experiments on three question-answering sets of MOOC Q\&A: 1-hop True/False, multi-hop True/False, and 1-hop Query.
The results are shown in Table \ref{table:main}. It is evident that the baseline methods, IO and CoT, which do not utilize external knowledge, exhibit very low accuracy rates in answering questions. This implies that these questions are not familiar to LLMs, and answering them requires a high degree of reliance on knowledge within the knowledge graph.

On the True/False sets, FiSKE demonstrates outstanding performance, surpassing the previous state-of-the-art methods. In 1-hop questions, even the performance of FiSKE based on GLM4-9B-Chat surpasses baseline methods based on GPT-4. In multi-hop questions, FiSKE still demonstrates significant advantages. This demonstrates the significant advantages of the new paradigm we propose over existing paradigms. FiSKE showed improvements when different models such as GLM4-9B-Chat, GPT-3.5, and GPT-4 were used. This indicates that FiSKE is very friendly to LLMs with fewer parameters, enabling better results at a lower cost, while still offering considerable room for improvement with better models.

On the Query set, considering the partial match metric, in the case of GPT4, results of ToG and PoG are only slightly worse than FiSKE. However, when based on GPT-3.5 and GLM4-9B-Chat, ToG and PoG perform significantly worse than FiSKE. This once again highlights the usability of our approach with LLMs that have smaller parameter sizes. Examining the complete match metric, our method consistently outperforms existing methods across all scenarios. This indicates that while existing methods can identify partial answers to questions, it struggles to comprehensively retrieve all possible answers to a given question. In contrast, our method excels in providing a comprehensive set of answers to the posed questions. 

The results of WebQuestions and WebQSP are shown in Table \ref{table:freebase}.
From the baseline, we can observe that LLMs can answer a considerable portion of the questions solely based on their own knowledge, indicating that these questions are relatively familiar to LLMs. To better illustrate whether these methods answer questions based on the inherent knowledge of LLMs or the retrieved knowledge, we merged the responses of these methods with the IO answers of LLMs and conducted additional comparisons. These are respectively represented as Method + IO. 

It can be observed that when dealing with a large-scale external knowledge base like Freebase, our proposed FiSKE still maintains excellent performance, generally outperforming existing methods across the board. FiSKE consistently demonstrates superior performance under both partial and complete matching evaluation criteria, and FiSKE + IO demonstrates potential for substantial further improvement. This indicates that FiSKE's significant advantage over existing methods stems primarily from its more accurate retrieval of knowledge from the knowledge graph.

The results of the agricultural graph are shown in Table \ref{table:agr}. It can be seen that the method we proposed outperforms ToG on single-hop questions, but does not show an advantage on multi-hop questions. The agricultural knowledge graph we constructed exhibits high density, where the same set of entities can form different pathways through diverse relational connections. FiSKE's suboptimal performance on multi-hop QA tasks likely stems from its current capability to only identify that all entities along these pathways are relevant to the clue, while lacking sufficient discriminative information to make precise selections among multiple candidate pathways. In our future work, we plan to extract richer contextual signals from the input questions to facilitate more precise path selection in the knowledge graph.

In summary, FiSKE has obvious advantages over existing methods, with significantly higher accuracy and completeness of answers compared to existing methods. However, when encountering overly specialized and complex questions, FiSKE performs on par with existing methods.

\subsection{Computational Cost}

In the knowledge base question-answering process, the majority of computational resources are dedicated to utilizing the Large Language Model, while the remaining computational consumption is relatively negligible. We calculate the average number of LLM calls during the question-answering process to illustrate the computational resource consumption. The results are shown in Table \ref{tab:llm_call}. 

We conducted experiments across different datasets and various base models. The results demonstrate that different question types significantly affect the number of LLM calls, while different base models also cause fluctuations in call frequency. However, overall, PoG requires significantly fewer LLM calls than ToG, and FiSKE requires even fewer calls than PoG. This clearly demonstrates FiSKE's superior efficiency in knowledge retrieval, and its high performance can effectively facilitate the practical application of question-answering methods.
\begin{table*}[htb]
    \renewcommand{\arraystretch}{1}
    \caption{Average Number of Calls to LLMs On MOOC Q\&A. }
    \centering
    \resizebox{\linewidth}{!}{
    \begin{tabular}{l|c|c|c}
    \hline
        \multicolumn{1}{c|}{\textbf{Method}} & \textbf{1-Hop True/False} & \textbf{Multi-Hop True/False}  & \textbf{1-Hop Query} \\
        \hline
        ToG \cite{sun2024think} w/GLM4-9B-Chat & 14.76 & 16.82 & 8.94\\
        ToG \cite{sun2024think} w/GPT-3.5      &19.81  & 22.16 & 11.20\\
        ToG \cite{sun2024think} w/GPT-4 &22.60  & 26.17 & 8.06\\
        \hline
        PoG \cite{chen2024plan} w/GLM4-9B-Chat & 11.87 & 14.89 & 7.62 \\
        PoG \cite{chen2024plan} w/GPT-3.5     & 14.53 & 15.71 & 7.29 \\
        PoG \cite{chen2024plan} w/GPT-4       & 11.21 & 13.14 & 6.82 \\
        \hline
        
        FiSKE w/GLM4-9B-Chat & 6.64 & 9.32 & \textbf{4.71}  \\
        FiSKE w/GPT-3.5      & 5.86 & 8.18 & 5.58\\
        FiSKE w/GPT-4 & \textbf{5.19} & \textbf{6.93} & 4.97\\
        \hline
    \end{tabular}
    }
    \label{tab:llm_call}
\end{table*}

\subsection{Studies on Clue Extraction}
\label{sce}
We evaluate the impact of clue extraction on the overall method. Four different clue variants were established in our study, as shown in Table \ref{tab:clue}.
(a) SP-Matched: Only clues that directly match a starting point in the knowledge graph are retained.
(b) Clue-Trunc: We randomly select one clue (not starting point) to modify by deleting the original clue, then keeping only its first word (for multi-word clues) or repeating it (for single-word clues). Other clues remain unchanged.
(c) Clue-Ext: Retains all original clues while adding their truncated variants to one randomly selected clue (same modification rules as (b)).
(d) Noise-Add: Introduces irrelevant clues to measure resistance to noise.

Approximately 21\% of QA pairs failed to locate starting points in the knowledge graph. For these cases, FiSKE defaults to CoT reasoning. When excluding these problematic pairs (as shown in SP-Matched results), accuracy improves at the cost of significantly higher computational overhead and processing time.
The Clue-Trunc variant demonstrates reduced accuracy and computational costs because truncated clues often prematurely terminate knowledge exploration when failing to find valid paths. In Clue-Ext experiments, the additional clues increase exploration cycles, leading to substantially higher computational overhead while paradoxically decreasing accuracy.
The irrelevant clues trigger unnecessary exploration cycles in the knowledge graph, all of which fail immediately due to semantic mismatch. While this still incurs computational overhead, the costs remain slightly lower than Clue-Ext because failed explorations terminate faster. All unsuccessful queries automatically default to CoT responses, resulting in accuracy identical to pure CoT performance.

\begin{table*}[htbp]
\centering
\caption{Experimental Results of Clue Extraction Variants on WebQSP (Llama3-8B-Instruct).}
    \resizebox{0.9\linewidth}{!}{
\label{tab:clue}
\begin{tabular}{llcccccc}
\toprule
\textbf{Dataset} & \textbf{Method}  & \multicolumn{1}{l}{\textbf{Partial Match}} & \multicolumn{1}{l}{\textbf{Complete Match}}& \textbf{LLM Call} & \multicolumn{1}{l}{\textbf{Total Token}}& \multicolumn{1}{l}{\textbf{LLM Time}} & \multicolumn{1}{l}{\textbf{Total Time}}   \\
\midrule

\multirow{3}{*}{WebQSP} & FiSKE & 70.8 & 40.4 & 6.9 & 3814.2  &49.5 & 57.7\\
 & SP-Matched & 72.6 & 41.9 & 8.4 & 4819.1  &61.6&72.0\\
 & Clue-Trunc & 67.4 & 39.1 & 5.4 & 2583.0  &31.1&39.0\\
 & Clue-Ext& 64.1& 36.2 & 8.4 & 4664.3&60.9&73.1  \\
  & Noise-Add & 62.1& 33.7 & 7.9 & 4263.0&60.2&77.5  \\

\bottomrule
\end{tabular}
}
\end{table*}

\begin{table}[htb]
    \renewcommand{\arraystretch}{1}
    \caption{Results for different backbone LLMs on WebQSP. Hybrid denotes a configuration where clue extraction is performed by GPT-3.5 while adaptive mapping strategy is executed on Llama3-8B-Instruct.}
    \centering
    \resizebox{\linewidth}{!}{
    \begin{tabular}{l|c|c}
    \hline
        \multicolumn{1}{c|}{\textbf{Method}} & \textbf{Partial Match} & \textbf{Complete Match}  \\
        \hline
        FiSKE w/GPT-3.5 & 83.1 & 49.1 \\
        FiSKE w/Hybrid     & 80.6 & 46.5\\
        FiSKE w/Llama3-8B-Instruct & 70.8  & 40.4\\
        \hline
    \end{tabular}
    }
    \label{tab:diff_llm}
\end{table}

\subsection{Studies on Different Backbone LLMs}
We conducted experiments on how FiSKE's performance varies when implementing its Clue Extraction and Adaptive Mapping Strategy (AMS) modules with different LLMs, as shown in Table \ref{tab:diff_llm}. When replacing the LLM used in the AMS phase from GPT-3.5 to Llama3-8B-Instruct, FiSKE's performance showed a slight decline, indicating that AMS has relatively low requirements for LLM capability. However, when substituting the LLM used in the clue extraction phase from GPT-3.5 to Llama3-8B-Instruct, FiSKE's performance decreased significantly. This primarily occurs because the correctness of clue extraction substantially impacts the method's effectiveness, as further demonstrated in Section \ref{sce}.

\begin{table*}[t]
\centering
\caption{Ablation Studies on WebQuestions set. SR represents the Stateful Record, AMS represents the Adaptive Mapping Strategy, and BAA represents Branch-Adaptive Answering.}
\resizebox{1\linewidth}{!}{
\begin{tabular}{lcccccc}
\toprule
\textbf{Variant} & \textbf{SR} & \textbf{AMS} &\textbf{BAA} & \textbf{partial match}& \textbf{complete match}& \textbf{avg LLM Calls} \\
\midrule
FiSKE       & \checkmark & \checkmark   & \checkmark   & 70.3 & 44.6 & 8.76        \\
\hline
w/o BAA            & \checkmark & \checkmark & \texttimes     &58.4 & 35.6   & 8.76\\
\hline
w/o SR   & \texttimes     & \checkmark & \checkmark    &63.7 &  42.7    & 5.94            \\
w/o SR+BAA   & \texttimes     & \checkmark     & \texttimes     &48.3 &  31.7 & 5.94        \\
\hline
w/o AMS            & \checkmark  & \texttimes  & \checkmark     &65.4 &  41.3 & 5.57 \\
w/o AMS+BAA           & \checkmark    & \texttimes    & \texttimes     &42.6 &  24.1 & 5.57\\
\hline
w/o SR+AMS            & \texttimes & \texttimes  & \checkmark     &61.2& 40.0  & 6.24\\
w/o SR+AMS+BAA           & \texttimes    & \texttimes    & \texttimes     &41.9 &23.8 & 6.24\\

\bottomrule
\end{tabular}
}
\label{tab:ablation}
\end{table*}

\subsection{Ablation Studies}

We conducted ablation studies and the results are shown in Table \ref{tab:ablation}.
We conducted ablation experiments on the three innovative components of FiSKE. In the Table \ref{tab:ablation}, SR stands for Stateful Record, AMS represents the Adaptive Mapping Strategy, and BAA denotes Branch-Adaptive Answering. Specifically, when removing the Stateful Record, we still conducted experiments based on fine-grained clues but no longer recorded the mappings between clues and entities or relationships in the knowledge graph. When removing the Adaptive Mapping Strategy, we still performed relation mapping exploration and entity mapping exploration but enforced their mapping to the same clue rather than adaptively selecting one or multiple clues. When removing Branch-Adaptive Answering, we no longer determined whether to leverage the LLM's own knowledge but uniformly passed the retrieved triples to the LLM. We not only tested the effects of removing each component individually but also experimented with the results of removing their combinations.

Overall, we can observe that each new component of FiSKE contributes to improving the overall performance of the method. Notably, Branch-Adaptive Answering has a particularly significant impact on the results. We argue that when the retrieved results are incomplete or even incorrect, directly feeding them to the LLM can substantially hinder its ability to correctly utilize its own knowledge for answering—especially for smaller-scale LLMs. Therefore, our strategy only provides the retrieved knowledge to the LLM after successfully completing all clue mappings, effectively mitigating this negative effect.

Another observation is that removing either Stateful Record or Adaptive Mapping Strategy reduces the number of LLM calls. Our analysis suggests that without these components, FiSKE tends to prematurely terminate exploration and instead directly query the LLM. While this decreases invocation frequency, it undermines effective knowledge retrieval from the knowledge graph. Consequently, when faced with questions beyond the LLM's capability, performance degrades significantly.

Finally, there is an implicit yet noteworthy observation: Even when we simultaneously remove both Stateful Record and Adaptive Mapping Strategy—effectively reducing FiSKE's knowledge exploration process to a level comparable to ToG—its performance still surpasses ToG. We attribute this advantage to our retention of fine-grained clues. Unlike ToG, which conducts knowledge exploration at the entire-question level, targeting fine-grained clues leads to more precise exploration results.

\section{Conclusion}
In this paper, we propose a new paradigm of fine-grained stateful exploration called FiSKE. It aims to address the widespread issue of knowledge granularity mismatch in existing knowledge graph question-answering paradigms. We analyze that this problem leads to redundant exploration or omission of important information during the knowledge exploration process, ultimately resulting in increased computational consumption and degraded question-answering performance. After decomposing questions into fine-grained clues, FiSKE maintains a stateful record to avoid redundant exploration or omissions, employs an adaptive mapping strategy to resolve ambiguity in clue-to-graph mappings, and automatically determines how explored knowledge paths are used for LLM augmentation. We conducted experiments based on three different knowledge graphs and corresponding multiple question-answering sets, and the results validated our design concepts. Our method significantly outperforms existing methods in terms of both question-answering results and computational efficiency. Additionally, the experiments demonstrate that our method still performs well with LLMs that have fewer parameters, showcasing great practical potential. Ablation studies also prove the effectiveness of each component of our design.

\appendix
\section{Appendix}
\subsection{Prompt Templates}
\label{appendix prompt}
\begin{figure*}
    \centering
    \includegraphics[width=0.8\linewidth]{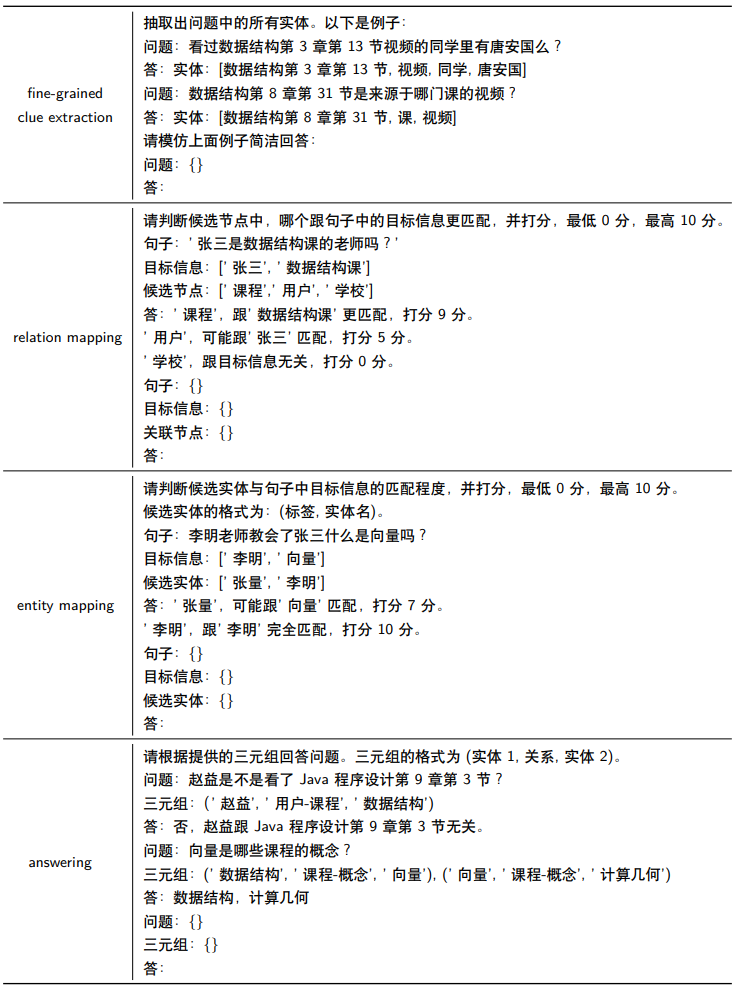}
    \caption{LLM prompts for Chinese datasets.}
    \label{fig:cn_prompts}
\end{figure*}

Our method employs four prompt templates to guide the LLM in performing four different tasks. Example prompts are shown in Table \ref{tab:prompts} and Fig. \ref{fig:cn_prompts}. FiSKE is not highly sensitive to the quality of the prompts. Our experiments include both Chinese and English datasets. We simply translated the prompts and modified the examples within them, without making any further specialized adjustments. For datasets in the same language, we used the same set of prompts. In Section \ref{fine}, we use the fine-grained clue extraction prompt to obtain fine-grained clues. In Section \ref{ke}, we apply the relation mapping prompt and the entity mapping prompt to complete the two steps of mapping exploration. Finally, in Section \ref{answering}, we utilize the answering prompt to instruct the LLM to answer questions based on the provided knowledge.

\begin{table*}

    \centering
    \caption{Examples of all LLM prompts used in this paper. These are designed for the Freebase knowledge base and can be appropriately adapted for other knowledge bases.}
    \resizebox{1\linewidth}{!}{
    \begin{tabular}{l|l}
    \toprule
        \makecell{fine-grained\\ clue extraction} & \makecell[l]{Please identify the entities in the sentence by taking into account the sentence's syntactic structure. \\
         Below is an example.\\
sentence: Which place is the madam satan located?\\
entities: ['place', 'madam satan']\\
Now answer with the format of the example above.\\
sentence: \{\} }\\
\midrule

        \makecell{relation mapping} & \makecell[l]{Please retrieve relations that are related to any of the target information from the sentence and rate it on a scale of 0 to 10. \\ Be brief and concise. \\
sentence: Name the president of the country whose main spoken language was Brahui in 1980?\\
target information: ['president', 'country', 'main spoken language', 'Brahui', '1980']\\
relations: ['language.human\_language.main\_country', \\'language.human\_language.human\_language', \\'language.human\_language.language\_family', \\'language.human\_language.iso\_639\_3\_code', \\'base.rosetta.languoid.parent', \\'language.human\_language.writing\_system', \\'base.rosetta.languoid.languoid\_class'] \\
assess: 
- 'language.human\_language.main\_country' is related to the country, so it's a match, (9) score. \\
- 'language.human\_language.human\_language' is kind of related to the main spoken language, so it's a match, (6) score.\\
- other relations are not related to any information in the sentence, so they are not match, (0) score.\\
sentence: \{\}
\\target information: \{\}
\\relations: \{\}
\\assess: }\\
\midrule
        \makecell{entity mapping} & \makecell[l]{Please assess the relevance of each candidate entity to the information in the sentence and assign a score ranging from 0 to 10. \\The format of the candidate entity is (relation, entity).\\
sentence: Which place is the madam satan located?\\
information in the sentence: ['place', 'madam satan']\\
candidate entity: [('film.film.country', 'the USA'), ('film.film.language', 'English')]\\
assess: 
- ('film.film.country', 'the USA') is related to 'place', so it's a match, (8) score.\\
- ('film.film.language', 'English')is not related to any information in the sentence,(0) score.\\
sentence: \{\}\\
information in the sentence: \{\}\\
candidate entity: \{\}\\
assess:}\\
         \midrule
        \makecell{answering} & \makecell[l]{please answer the question based on the provided triplets. The format of the triplet is (entity1, relation, entity2). \\If there is not enough information provided in the triplets, answer the question with your own knowledge.\\
question: Which place is the madam satan located?\\
triplets: [('madam satan', 'film.film.country', 'the USA'), ('madam satan', 'film.film.language', 'English')]\\
answer: 'madam satan' is located in 'the USA'.\\
question: \{\}\\
triplets: \{\}\\
answer: }\\
\bottomrule
         
    \end{tabular}
    }
    \label{tab:prompts}
\end{table*}

\subsection{Pseudocode}
\label{pseudocode}
The pseudocode of Adaptive and Stateful Knowledge Exploration can be seen in Algorithm \ref{alg:knowledge_exploration}.
\begin{algorithm}
\caption{Pseudocode of Adaptive and Stateful Knowledge Exploration}
\label{alg:knowledge_exploration}
\DontPrintSemicolon
\KwIn{Initial entity $e^0$, all clues $clue_{all}$, knowledge graph $\mathcal{G}$}
\KwOut{Knowledge path $P$, explored clues $C$}
Initialize $P \leftarrow \emptyset$, $C \leftarrow \emptyset$, $n \leftarrow 1$\;
$C' \leftarrow clue_{all} \setminus C$\; \tcp*{Candidate clues}
\While{not terminated}{
    \tcp{Relation-Based Mapping (Step $n$)}
    $R \leftarrow \text{GetRelations}(\mathcal{G}, e^{n-1})$\; 
    $\text{r-mapping} \leftarrow \text{LLM.RMapping}(C', R)$\;
    
    \If{$\text{r-mapping} = \emptyset$}{
        Terminate\;
    }
    
    \tcp{Entity-Based Mapping (Step $n$)}
    \ForEach{$(\text{clue}_i, r_j) \in \text{r-mapping}$}{
        $E_j \leftarrow \text{GetEntities}(\mathcal{G}, e^{n-1}, r_j)$\; 
    }
    $\text{e-mapping} \leftarrow \text{LLM.EMapping}(C', \bigcup_j E_j)$\;
    
    \eIf{$\text{e-mapping} = \emptyset$}{
        Terminate\; \tcp*{Case 3: No entity mapping}
    }{
        \tcp{Handle successful mappings}
        \If{$\exists (\text{clue}_1, r_i), (\text{clue}_1, e_j)$}{
            \tcp*{Case 1: Single-clue dual mapping}
            $C \leftarrow C \cup \{\text{clue}_1\}$\;
            $P \leftarrow P \cup \{(e^{n-1}, r_i, e_j)\}$\;
            $e^{n} \leftarrow e_j$\;
        }
        \ElseIf{$\exists (\text{clue}_1, r_i), (\text{clue}_2, e_j)$}{
            \tcp*{Case 2: Dual-clue mapping}
            $C \leftarrow C \cup \{\text{clue}_1, \text{clue}_2\}$\;
            $P \leftarrow P \cup \{(e^{n-1}, r_i, e_j)\}$\;
            $e^{n} \leftarrow e_j$\;
        }
        $n \leftarrow n + 1$\;
        $C' \leftarrow clue_{all} \setminus C$\; 
    }
}
\Return $P, C$\;
\end{algorithm}
\printcredits

\bibliographystyle{cas-model2-names}

\bibliography{cas-refs}





\end{document}